\pdfoutput=1
\documentclass[letterpaper, 10 pt, conference]{ieeeconf}
\IEEEoverridecommandlockouts    %
\usepackage{hyperref}
\usepackage{threeparttable}
\usepackage{algorithm}
\usepackage{cite}
\usepackage{pifont}
\usepackage{ragged2e}
\hypersetup{
    colorlinks=true,
    linkcolor=magenta,
    filecolor=magenta,      
    urlcolor=magenta,
}

\usepackage{wrapfig}
\usepackage{lipsum}
\usepackage{xspace}
\newcommand{\etal}{\emph{et al.}\xspace}
\usepackage{multirow}
\usepackage[table]{xcolor} %
\definecolor{lightgray}{RGB}{220, 220, 220} %
\usepackage{caption}
\usepackage{subcaption}

\usepackage{graphics}           
\usepackage{times}              
\usepackage{amsmath}            
\usepackage{amssymb}            
\usepackage{graphicx}
\usepackage{algorithm}
\usepackage[noend]{algpseudocode}
\usepackage{booktabs}
\usepackage{color}
\definecolor{instructioncolor}{rgb}{.5,.5,.5}

\usepackage[font=small]{caption}

\def\eqref#1{Eq.~(\ref{#1})}

\makeatletter
\usepackage{xspace}
\DeclareRobustCommand\onedot{\futurelet\@let@token\@onedot}
\def\@onedot{\ifx\@let@token.\else.\null\fi\xspace}

\def\etal{{et al}\onedot}
\makeatother

\usepackage{array}
\newcolumntype{L}[1]{>{\raggedright\let\newline\\\arraybackslash\hspace{0pt}}m{#1}}
\newcolumntype{C}[1]{>{\centering\let\newline\\\arraybackslash\hspace{0pt}}m{#1}}
\newcolumntype{R}[1]{>{\raggedleft\let\newline\\\arraybackslash\hspace{0pt}}m{#1}}

\usepackage{bm}
\usepackage{multirow}
\usepackage{rotating}  %

\renewcommand{\thefigure}{\Roman{figure}}
\newcommand{\myblue}[1]{{\color{blue}#1}}
\definecolor{mygreen_rgb}{RGB}{0,0,0}

\title{\LARGE \bf Novel Diffusion Models for Multimodal 3D Hand Trajectory Prediction}

\author{Junyi~Ma$^1$, Wentao~Bao$^2$, Jingyi~Xu$^3$, Guanzhong~Sun$^4$, Xieyuanli~Chen$^{5}$, Hesheng~Wang$^{1*}$
\thanks{$^{1}$Junyi~Ma and Hesheng~Wang are with IRMV Lab, the Department of Automation, Shanghai Jiao Tong University.}
\thanks{$^{2}$Wentao~Bao is with Meta Reality Labs.}
\thanks{$^{3}$Jingyi~Xu is with the Department of Electronic Engineering, Shanghai Jiao Tong University.}
\thanks{$^{4}$Guanzhong~Sun is with the School of Information and Control Engineering, China University of Mining and Technology.}
\thanks{$^{5}$Xieyuanli~Chen is with the College of Intelligence Science and Technology, National University of Defense Technology.}
\thanks{$^{*}$Corresponding author email: wanghesheng@sjtu.edu.cn}
}

\begin{document}
\maketitle

\IEEEpeerreviewmaketitle
\thispagestyle{empty}
\pagestyle{empty}

\begin{abstract}

Predicting hand motion is critical for understanding human intentions and bridging the action space between human movements and robot manipulations. Existing hand trajectory prediction (HTP) methods forecast the future hand waypoints in 3D space conditioned on past egocentric observations. However, such models are only designed to accommodate 2D egocentric video inputs. There is a lack of awareness of multimodal environmental information from both 2D and 3D observations, hindering the further improvement of 3D HTP performance.
In addition, these models overlook the synergy between hand movements and headset camera egomotion, either predicting hand trajectories in isolation or encoding egomotion only from past frames. To address these limitations, we propose novel diffusion models (MMTwin) for multimodal 3D hand trajectory prediction. MMTwin is designed to absorb multimodal information as input encompassing 2D RGB images, 3D point clouds, past hand waypoints, and text prompt. Besides, two latent diffusion models, the egomotion diffusion and the HTP diffusion as twins, are integrated into MMTwin to predict camera egomotion and future hand trajectories concurrently. We propose a novel hybrid Mamba-Transformer module as the denoising model of the HTP diffusion to better fuse multimodal features. The experimental results on three publicly available datasets and our self-recorded data demonstrate that our proposed MMTwin can predict plausible future 3D hand trajectories compared to the state-of-the-art baselines, and generalizes well to unseen environments. The code and pretrained models will be released at \url{https://github.com/IRMVLab/MMTwin}.

\end{abstract}

\section{Introduction}
\label{sec:intro}

Understanding how humans behave has become increasingly important in robot learning and extended reality. Although various algorithms have been proposed to recognize and anticipate coarse-grained action categories~\cite{bao2022opental,xu2023dynamic,qi2024uncertainty}, analyzing fine-grained hand motion closely associated with human behaviors has gradually gained attention. In the context where some works~\cite{Ye_2023_ICCV,zhang2024hoidiffusion,zhu2023get} focus on reconstructing hand grasping states with target objects,
how to achieve future hand trajectory prediction (HTP) in 2D and 3D spaces with egocentric vision remains a challenging problem. The high uncertainty of hand motion in first-person views determines the difficulty of fitting long-term hand waypoint distributions. 

\begin{figure}
  \centering
  \includegraphics[width=1\linewidth]{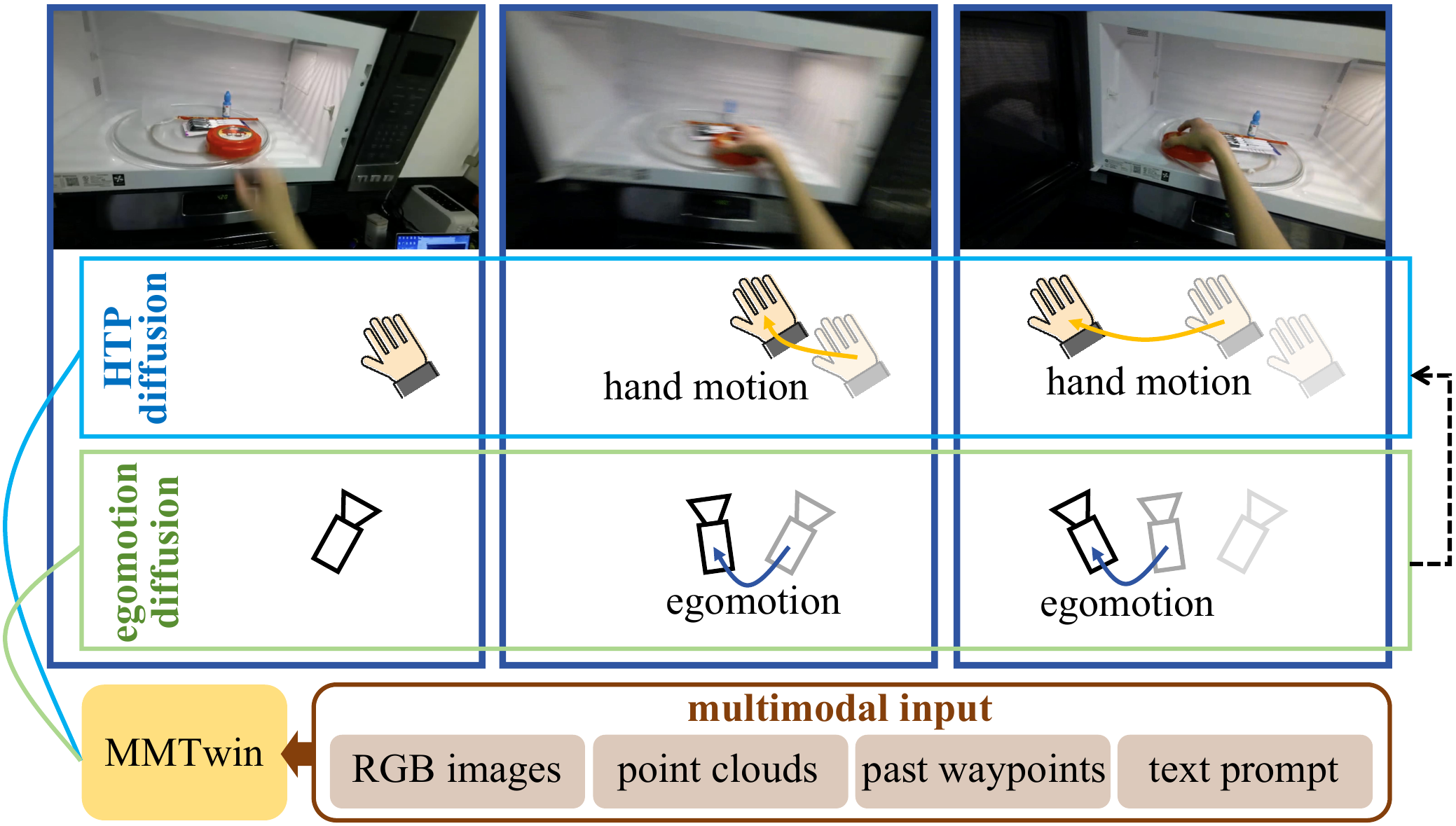}
  \caption{MMTwin receives multimodal data to concurrently predict future camera egomotion and hand trajectories with twin diffusion models. It attends to 3D structure awareness and synergy between hand movements and camera egomotion in future time periods.}
  \label{fig:motivation_maintext}
  \vspace{-0.7cm}
\end{figure}

Compared to the 2D HTP task, predicting hand waypoints in 3D space can be exploited for a wider range of applications such as robotic end-effector planning. However, the existing HTP models~\cite{bao2023uncertainty,liu2022joint,ma2024diff} are only designed to process 2D egocentric video inputs and overlook incorporating 3D structure awareness. Since humans use stereo vision to perceive 3D environmental features in any interaction process, a gap inevitably exists between predicted trajectories and real hand motion due to 2D-3D input modality discrepancies. This hinders further performance improvement in the literature of 3D HTP. In addition, humans move their hands as a part of their body according to their intentions, and thus comprehensively analyzing synergy body motion is essential for accurately predicting 3D hand trajectories. Although the recent work \cite{ma2024madiff} considers the effect of headset camera egomotion in the hand-related state transition process, it is still limited to analyzing these two coupled motions within the past time durations. As the sequential images in Fig.~\ref{fig:motivation_maintext}, the view of the headset camera turns to the left side and concurrently the right hand also moves the target object to the left side. There is an entanglement between hand movements and camera egomotion within both past and future interaction processes in egocentric views, which needs to be explicitly decoupled for better understanding and predicting hand motion.

In this work, we develop 3D HTP by incorporating comprehensive 2D and 3D observations for better environmental perception. Our unified HTP framework integrates multimodal inputs, including 2D RGB images, 3D point clouds, past hand waypoints, and text prompt. To decouple camera egomotion and hand motion predictions, we develop twin diffusion models, egomotion diffusion and HTP diffusion, as shown in Fig.~\ref{fig:motivation_maintext}. It explicitly captures the synergy by predicting future 3D hand waypoints conditioned on predicted egomotion features. To better harmonize multimodal features within the diffusion process, we propose a novel denoising model with a hybrid Mamba-Transformer architecture for diffusion models. In the devised hybrid pattern, voxel patches from the 3D input modality are fused with HTP latents by the structure-aware Transformer to capture 3D global context. Besides, camera egomotion features predicted by the egomotion diffusion are also integrated into the egomotion-aware Mamba for reasonable state transition in future time horizons. This combines Mamba’s strength in temporal modeling with Transformer’s ability to capture global context, improving multimodal 3D hand trajectory prediction.

The main contributions of this work are as follows: 
\begin{itemize}
    \item We propose novel \underline{twin} diffusion models dubbed MMTwin for 3D hand trajectory prediction, which exploits \underline{m}ulti\underline{m}odal information as input to concurrently predict future camera egomotion and hand movements in egocentric views.
    \item A hybrid Mamba-Transformer module is designed for the denoising model in the HTP diffusion to harmonize multimodal features. It fuses 3D global context by the structure-aware Transformer after the state transition of HTP latents in the egomotion-aware Mamba.
    \item The experimental results show that our proposed MMTwin can predict more plausible future 3D hand trajectories compared to the state-of-the-art (SOTA) baselines, and shows good generalization ability to unseen environments.
    
\end{itemize}

\vspace{-0.1cm}
\section{Related work}
\label{sec:related_work}

In recent years, the importance of HTP has grown significantly in extended reality and service robots, such as aiding patients with neuromuscular diseases by suggesting feasible future hand waypoints~\cite{bao2023uncertainty,li2022egocentric}. HTP also bridges human motion and robot manipulation by transferring hand prediction to end-effector planning~\cite{bahl2023affordances,mendonca2023structured,singh2024hand}. However, accurately forecasting hand waypoints in egocentric views remains challenging.
Here we review this literature according to whether the states of the target objects are explicitly perceived, introducing object-aware and object-agnostic hand trajectory prediction accordingly.

\textbf{Object-aware hand trajectory prediction.}
Human hand movements are typically performed purposefully around the target object~\cite{chen2024object,zhang2024hoidiffusion,ju2025robo} during hand-object interaction (HOI). Therefore, some prior works attend to jointly forecasting future hand trajectories and target object affordance in egocentric videos. They ensure the awareness of interacted object states when analyzing how hands move. Liu~\etal~\cite{liu2020forecasting} pioneer the concurrent prediction of hand motor attention and object affordances using a convolution-based backbone. In contrast, OCT~\cite{liu2022joint} uses an object-centric Transformer for autoregressive forecasting of future hand trajectories and interaction hotspots. More recently, Zhang~\etal~\cite{zhang2024pear} propose a multitask network to capture human intention as well as manipulation. Diff-IP2D~\cite{ma2024diff} first adopts a diffusion model to achieve HOI prediction on 2D egocentric videos. It forecasts future HOI latents which are further decoded by the devised heads to generate hand waypoint distributions and target object affordances. All these approaches predict future hand trajectories while keeping the awareness of target objects. There is always an over-reliance on the prior object position/feature extraction with the off-the-shelf detectors~\cite{shan2020understanding} or utilizing predefined object-related phrase~\cite{zhang2024pear}. 

\textbf{Object-agnostic hand trajectory prediction.}
To improve inference efficiency and robustness to multiple interaction environments, some recent works turn to object-agnostic HTP, which eliminate the need for prior object detection and verb-noun descriptions. These object-agnostic schemes align better with the trendy end-to-end manner in embodied intelligence.
For example, Bao \etal \cite{bao2023uncertainty} achieve 3D hand trajectory prediction in egocentric views by an uncertainty-aware state space Transformer in an autoregressive manner. Tang~\etal~\cite{tang2025prompting} predicts future 3D coordinates of multiple hand joints without observing target objects.
Gamage \etal~\cite{gamage2021so} design a hybrid classical-regressive kinematics model for structured and unstructured ballistic hand motion in VR activities. Recently, MADiff~\cite{ma2024madiff} is proposed to predict future 2D hand trajectories without explicitly detecting target objects, which instead uses a foundation model to extract environmental semantic features. In this work, we also follow the object-agnostic paradigm of MADiff~\cite{ma2024madiff}. Notably, we extend its 2D predictive task in egocentric views to 3D space, which enriches multimodal observations by 3D point cloud input and directly outputs future 3D hand waypoints. Then we decouple the predictions of headset camera egomotion and hand movements by twin diffusion models. Moreover, we strengthen the denoising model with a hybrid Mamba-Transformer architecture, to achieve reasonable HTP state transition as well as 3D global context awareness.

\begin{figure*}
  \centering
  \includegraphics[width=1\linewidth]{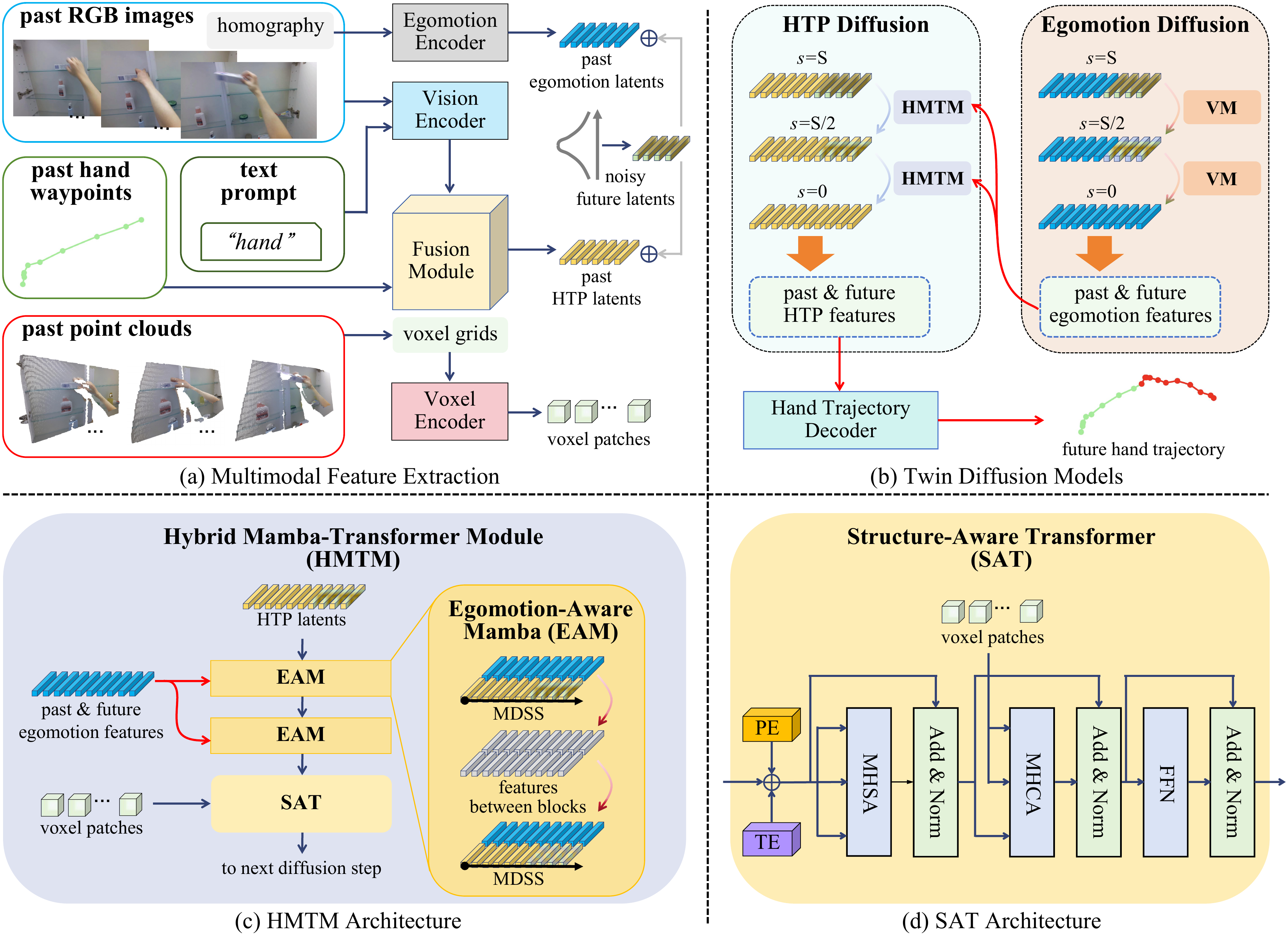}
  \caption{Our proposed MMTwin (a) extracts features from multimodal data, and (b) decouples predictions of future camera egomotion features and 3D hand trajectories by novel twin diffusion models. The vanilla Mamba (VM) is used for denoising in the egomotion diffusion. We further design a new denoising model in HTP diffusion with (c) a hybrid Mamba-Transformer module (HMTM), encompassing the egomotion-aware Mamba (EAM) blocks and (d) the structure-aware Transformer (SAT).}
  \label{fig:sytem_overview}
  \vspace{-0.5cm}
\end{figure*}

\section{Proposed Method}
\label{sec:proposed_method}

\subsection{System Overview}
\label{sec:overview}

Here we first provide the overall inference pipeline of our MMTwin in Fig.~\ref{fig:sytem_overview}. MMTwin receives multimodal data including egocentric RGB images $\mathcal{I}=\{I_t\}_{t=-N_\text{p}+1}^{0} \,(I_t \in \mathbb{R}^{c\times h\times w})$, point clouds $\mathcal{D}=\{D_t\}_{t=-N_\text{p}+1}^{0} \,(D_t \in \mathbb{R}^{n\times 3})$, past 3D hand waypoints $\mathcal{H}_p=\{H_t\}_{t=-N_\text{p}+1}^{0} \,(H_t \in \mathbb{R}^{3})$, and text prompt $\mathcal{O}$, and predicts future 3D hand trajectories $\mathcal{H}_\text{f}=\{H_t\}_{t=1}^{N_\text{f}}$. $N_\text{p}$ and $N_\text{f}$ denote the numbers of frames in the past and future time horizons respectively. $\mathcal{I}$ and $\mathcal{D}$ are both captured by headset RGBD camera. $\mathcal{O}$ is set as \textit{hand} as proposed in the previous work~\cite{ma2024madiff}. Following~\cite{liu2022joint,bao2023uncertainty}, we predict future hand waypoints in the global coordinate system, which is assigned as the camera coordinate system in the first frame of the input sequence $(t=-N_\text{p}+1)$. 

MMTwin first calculates sequential homography $\mathcal{M}=\{M_t\}_{t=-N_\text{p}+1}^{0} \,(M_t \in \mathbb{R}^{3\times 3})$ from $\mathcal{I}$ as past camera egomotion following~\cite{ma2024diff}, and encodes them to past egomotion latents for the egomotion diffusion. 
$M_t$ represents the homography matrix between $t\,\text{th}$ frame and the first frame $(t=-N_\text{p}+1)$ estimated by SIFT descriptors~\cite{lowe2004distinctive} with RANSAC \cite{fischler1981random}.
The egomotion diffusion predicts future egomotion latents conditioned on past ones, which will be further used as conditions for the HTP diffusion. The input images $\mathcal{I}$ and the text prompt $\mathcal{O}$ are fed into a foundation model~\cite{Li_2022_CVPR} to generate visual semantic features. A fusion module proposed by the previous work~\cite{ma2024madiff} incorporates past hand waypoints and visual semantic features to generate HTP latents for the HTP diffusion. The input sequential point clouds $\mathcal{D}$ are first transformed to the unified global coordinate system by visual odometry, and then we voxelize the aggregated points into discrete voxel grids to reduce memory consumption and improve running efficiency. A voxel encoder is built based on 3D convolutions to convert the dense grids to 3D voxel patches. We exploit the vanilla Mamba~\cite{gu2023mamba} (VM) as the denoising model in the egomotion diffusion, while we propose a hybrid Mamba-Transformer module (HMTM) for denoising in the HTP diffusion. The denoised HTP latents are ultimately decoded to future 3D hand waypoints $\mathcal{H}_\text{f}$. As can be seen, MMTwin achieves decoupling predictions of camera egomotion and 3D hand trajectories, and bridges them through denoising conditions, following the fact that there is a synergy between hand movements and camera egomotion within the future interaction process.

\begin{figure}[t]
  \centering
    \includegraphics[width=1\linewidth]{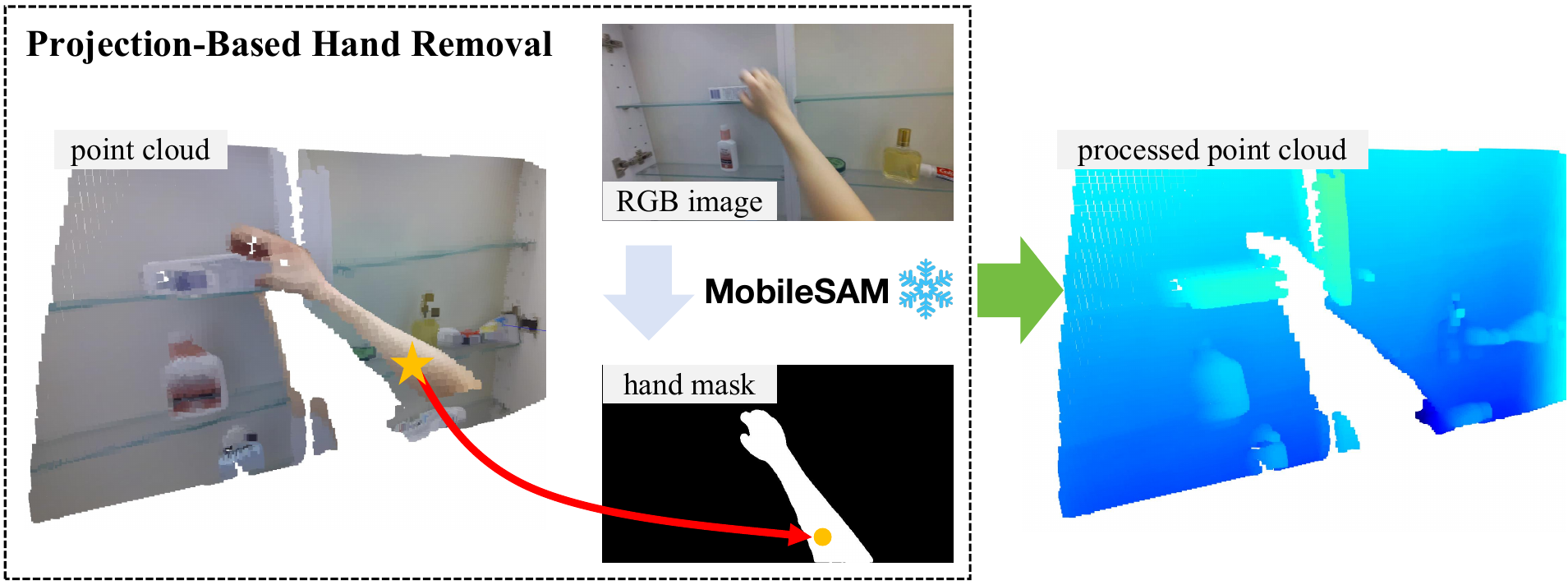}
  \caption{Projection-based hand removal. We use MobileSAM~\cite{mobile_sam} to generate the hand mask for each input image, and filter out the 3D points that are projected into the hand area by camera intrinsics.}
  \label{fig:hand_removal}
  \vspace{-0.7cm}
\end{figure}

\subsection{Multimodal Feature Extraction}

In this section, we provide detailed clarifications about how to transform the multimodal input data into the feature/latent spaces for the following twin diffusion models.

\textbf{Vision encoder.} As shown in Fig.~\ref{fig:sytem_overview}(a), we extract visual semantic features from past RGB images with text prompt \textit{hand} following the previous work~\cite{ma2024madiff}. We use pretrained GLIP~\cite{Li_2022_CVPR} here as the vision encoder. The visual grounding ability of GLIP enables
our visual encoder to semi-implicitly capture hand poses and
hand-scenario relationships within each 2D image frame, since the text prompt is used to indicate which part of the image should be focused on.
Specifically, we extract the outputs of the deepest cross-modality multi-head attention module (X-MHA) in GLIP, which are denoted as $X^\text{sem}\in \mathbb{R}^{(N_\text{p}+L)\times x}$. $x$ is the feature channel dimension. $L$ equals $N_\text{f}$ during training, and is set to $0$ during inference since future HTP latents will be replaced by sampled noises (noisy future latents in Fig.~\ref{fig:sytem_overview}(a)) in the inference process of our HTP diffusion. 

\textbf{Fusion module.} The fusion module first encodes past hand waypoints to trajectory features, and then uses $1\times 1$ convolution with Multilayer Perceptron (MLP) to fuse the trajectory features and the visual semantic features from the vision encoder. The output fusion features, denoted as $F^\text{htp}_\text{p}\in \mathbb{R}^{N_\text{p}\times f}$ and $F^\text{htp}_\text{f}\in \mathbb{R}^{N_\text{f}\times f}$, are regarded as HTP latents for our HTP diffusion. $f$ represents the channel dimension of HTP latents. $F^\text{htp}_\text{f}$ only exists in the training process for reconstruction supervision since noisy future latents $F^\text{htp}_\text{noise}\in \mathbb{R}^{N_\text{f}\times f}$ is concatenated to $F^\text{htp}_\text{p}$ in the inference process of our HTP diffusion.

\textbf{Voxel encoder.} Our proposed MMTwin achieves structure-aware 3D hand trajectory prediction by leveraging 3D perception. It is impractical to encode every input point cloud $D_t \in \mathcal{D}$ captured by the headset RGBD camera considering running efficiency and memory consumption. Therefore, we first transform them into the above-mentioned unified global coordinate system by visual odometry. Notably, for each frame, we use MobileSAM~\cite{mobile_sam} to remove point clouds projected to arms (as shown in Fig.~\ref{fig:hand_removal}). This is motivated by the fact that moving arms lead to cluttered points after multiple frame aggregation, which affect the precise representation of global 3D information. Then we voxelize them into voxel grids to avoid disturbance of unordered data structure, and further improve running efficiency and reduce memory consumption. Subsequently, we propose using the 3D-convolution-based voxel encoder to convert the dense 3D voxels into the sparse representation $X^\text{vox}\in \mathbb{R}^{N_\text{vox}\times f}$, which has $N_\text{vox}$ voxel patches with the same channel dimension $f$ as HTP latents. Note that in this work we do not integrate the voxel features into HTP latents because they are not time-varying due to the unified global representation. Instead, we advocate using them as the global context of the 3D interaction environments for the following denoising process in the HTP diffusion, which will be introduced in Sec.~\ref{sec:twin}.

\textbf{Egomotion encoder.} Following the previous work~\cite{ma2024diff,ma2024madiff}, we use one MLP to encode sequential homography matrices $\mathcal{M}$ into past egomotion features $F^\text{ego}_\text{p}\in \mathbb{R}^{N_\text{p}\times f}$ as latents for the egomotion diffusion. Similar to the setup of HTP latents, ground-truth future 
egomotion latents $F^\text{ego}_\text{f}\in \mathbb{R}^{N_\text{f}\times f}$ only exist in the training process for supervision and will be replaced by sampled noises $F^\text{ego}_\text{noise}\in \mathbb{R}^{N_\text{f}\times f}$ to enable denoising-based inference.

\begin{figure}[t]
  \centering
    \includegraphics[width=1\linewidth]{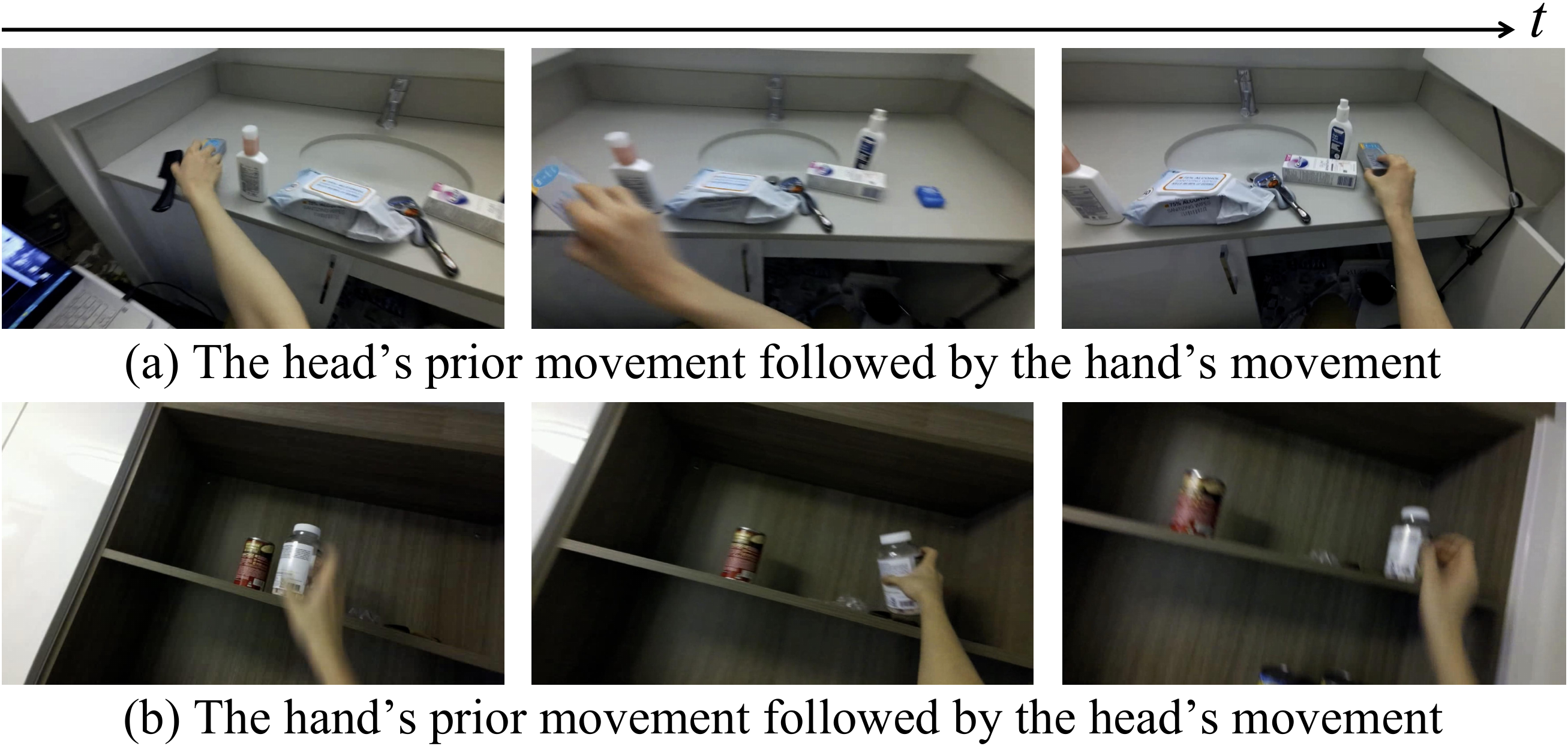}
  \caption{The exampled head movement (corresponding to camera egomotion) and hand movement coupled during the hand-object interaction process in egocentric views in the EgoPAT3D dataset~\cite{li2022egocentric}.}
  \label{fig:coupled_motion}
  \vspace{-0.7cm}
\end{figure}

\vspace{-0.1cm}
\subsection{Twin Diffusion Models}
\label{sec:twin}

The synergy between hand movements and headset camera egomotion within the future interaction process is reflected in three aspects: (1) human hand movements follow head movements in most cases as the head’s prior motion can provide valuable target observations for hand trajectory planning (Fig.~\ref{fig:coupled_motion}(a)), (2) head movements may conversely follow hand movements because hand actions sometimes occur subconsciously and are faster than head movements (Fig.~\ref{fig:coupled_motion}(b)), and (3) humans always aim to keep moving hands within their field of egocentric views to ensure accurate contact with the target object (Fig.~\ref{fig:coupled_motion}(a) and Fig.~\ref{fig:coupled_motion}(b)). We argue that predicting hand motion agnostic to future head motion does not align with real human behavior planning. Instead, explicitly decoupling the entangled head-hand movements helps 3D HTP models to better understand synergy motion patterns and potential intentions of interaction. Therefore, we propose novel twin diffusion models, i.e., the egomotion diffusion and HTP diffusion to predict headset camera egomotion and future hand trajectories concurrently.

\textbf{Egomotion diffusion.} As shown in Fig.~\ref{fig:sytem_overview}(b), the egomotion diffusion first converts noisy future egomotion latents $F^\text{ego}_\text{noise}$ to future egomotion homography features $\hat{F}^\text{ego}_\text{f}\in \mathbb{R}^{N_\text{f}\times f}$, which will be used as conditions for HTP diffusion. Here we leverage the vanilla Mamba~\cite{gu2023mamba} as the denoising model for efficient temporal modeling. The effect of multimodal data on the denoising model of the egomotion diffusion is achieved through gradient updates from narrowing $\hat{F}^\text{ego}_\text{f}$ and $F^\text{ego}_\text{f}$, as well as reducing trajectory prediction losses, rather than explicitly incorporating relevant multimodal features as denoising conditions. This is because head movement patterns are much simpler than those of moving hands. Here we omit the process of decoding the future homography features into specific homography matrices. We thus avoid uncertainties in selecting different supervision signals for possible homography formats~\cite{detone2016deep}.

\textbf{HTP diffusion.} 
As shown in Fig.~\ref{fig:sytem_overview}(b), our HTP diffusion takes in past HTP latents $F^\text{htp}_\text{p}$ to predict future counterparts $\hat{F}^\text{htp}_\text{f}$, conditioned on $F^\text{ego}_\text{p}\in \mathbb{R}^{N_\text{p}\times f}$ and $\hat{F}^\text{ego}_\text{f}$ predicted by the egomotion diffusion. Here we propose a novel hybrid Mamba-Transformer module (HMTM) as the denoising model. The architecture of HMTM is illustrated in Fig.~\ref{fig:sytem_overview}(c), which consists of two egomotion-aware Mamba (EAM) blocks and the structure-aware Transformer (SAT). EAM is first proposed by the recent work MADiff~\cite{ma2024madiff}, which designs motion-driven selective scan (MDSS) to seamlessly integrate egomotion homography features into the state transition process of Mamba. In this work, we first concatenate $F^\text{ego}_\text{p}$ with the predicted $\hat{F}^\text{ego}_\text{f}$ to $\hat{F}^\text{ego}_\text{pf} \in \mathbb{R}^{(N_\text{p}+N_\text{f})\times f}$, as well as $F^\text{htp}_\text{p}$ with the sampled noise $F^\text{htp}_\text{noise}$ to $F^\text{htp}_\text{pf} \in \mathbb{R}^{(N_\text{p}+N_\text{f})\times f}$, both along the time dimension. Then we implement MDSS in EAM for each denoising step of the HTP diffusion, to denoise the future part of $F^\text{htp}_\text{pf}$ conditioned on the holistic sequential egomotion feature $\hat{F}^\text{ego}_\text{pf}$. 
We refer more details of MDSS to the previous work~\cite{ma2024madiff}. We stack two EAM blocks here which are determined by the ablation in Sec.~\ref{sec:exp_albation}. 

Following stacked EAM blocks, the structure-aware Transformer is proposed in HMTM for each denoising step to capture 3D global context of the interaction environments for hand trajectory prediction. SAT helps to better fuse multimodal features from 2D and 3D observations. As shown in Fig.~\ref{fig:sytem_overview}(d), we first perform multi-head self-attention (MHSA) on HTP latents following positional/temporal encoding (PE/TE), and then implement multi-head cross-attention (MHCA) between sparse voxel features $X^\text{vox}$ and the output of MHSA, leading to the latents for the next diffusion step. Ultimately, we derive the denoised HTP features $\hat{F}^\text{htp}_\text{pf}$ after the last denoising diffusion step, of which the future part $\hat{F}^\text{htp}_\text{f}\in \mathbb{R}^{N_\text{f}\times f}$ is decoded by the MLP-based hand trajectory decoder (as shown in Fig.~\ref{fig:sytem_overview}(b)) to future 3D hand waypoints $\mathcal{H}_\text{f}$. As human perceives 3D environments with stereo vision to understand 3D global context such as spatial layout and collision information, MMTwin leverages voxel features from 3D point clouds as environmental global context for more reasonable 3D hand trajectory prediction. The hybrid pattern of Mamba and Transformer in HMTM is designed according to the ablation in Sec.~\ref{sec:exp_albation}.

\subsection{Traning and Inference}
\label{sec:training_and_inf}
Partial noising and denoising \cite{gong2022diffuseq} is adopted for the training and inference stages of both egomotion diffusion and HTP diffusion. That is, we anchor the past latents $F^\text{ego}_\text{p}$ and $F^\text{htp}_\text{p}$ in forward and reverse steps. To train MMTwin,
we use the diffusion-related losses $\mathcal{L}^\text{ego}_\text{VLB}$ for recovering future egomotion features in the egomotion diffusion, and use diffusion-related losses $\mathcal{L}^\text{htp}_\text{VLB}$, trajectory displacement loss $\mathcal{L}_\text{dis}$, regularization term $\mathcal{L}_\text{reg}$, and trajectory angle loss $\mathcal{L}_\text{angle}$ proposed in the prior works \cite{ma2024diff,ma2024madiff} for the HTP diffusion. 
The total loss function for MMTwin is the weighted sum of all the above-mentioned losses. We refer more details of the utilized loss functions to the previous works~\cite{ma2024madiff, ma2024diff}.

In the inference stage shown in Fig.~\ref{fig:sytem_overview}(b), we first denoise $F^\text{ego}_\text{noise}$ to $\hat{F}^\text{ego}_\text{f}$ in the egomotion diffusion, and then concatenate $F^\text{ego}_\text{p}$ with the predicted $\hat{F}^\text{ego}_\text{f}$ to $\hat{F}^\text{ego}_\text{pf}$. Next, $\hat{F}^\text{ego}_\text{pf}$ as the condition is fed to the HTP diffusion, which achieves denoising $F^\text{htp}_\text{noise}$ to $\hat{F}^\text{htp}_\text{f}$. Ultimately, $\hat{F}^\text{htp}_\text{f}$ is decoded to future hand waypoints by the hand trajectory decoder.

\begin{table*}[t]
\setlength{\tabcolsep}{14.35pt}
\center
\renewcommand\arraystretch{0.7}
\caption{Comparison of performance on hand trajectory prediction on the EgoPAT3D-DT and H2O-PT datasets in the 3D/2D space. Best and secondary results are viewed in \textbf{bold black} and \myblue{blue} colors respectively.}
\begin{tabular}{l|cc|cc|cc}
\toprule
\multicolumn{1}{l|}{\multirow{2}{*}{Approach}}   & \multicolumn{2}{c|}{EgoPAT3D-DT (seen)} & \multicolumn{2}{c|}{EgoPAT3D-DT (unseen)}  & \multicolumn{2}{c}{H2O-PT} \\ \cmidrule{2-7} 
\multicolumn{1}{c|}{}                                                                               & ADE\,$\downarrow$    & FDE\,$\downarrow$ & ADE\,$\downarrow$   & FDE\,$\downarrow$ & ADE\,$\downarrow$   & FDE\,$\downarrow$   \\ \cmidrule{1-7}                 
CVH \cite{ma2024diff}  & 1.100/0.221  	& 1.278/0.262  & 0.952/0.219   & 1.018/0.251  & 0.146/0.187    & 0.148/0.189 \\
OCT$^{*}$ \cite{liu2022joint}   &0.370/0.202    & 0.524/0.315  &  0.309/0.150   & 0.397/0.189  & 0.103/0.137  &0.126/0.152 \\ 
USST$^*$ \cite{bao2023uncertainty}  & \myblue{0.183}/0.089	  & \myblue{0.341}/0.274	 & \myblue{0.120}/0.075  & \textbf{0.185}/0.127	&\myblue{0.031}/\textbf{0.037}	 & \myblue{0.052}/\myblue{0.043} \\ 
S-Mamba \cite{wang2024mamba} & 0.185/0.084    & 0.355/0.141	&0.138/\myblue{0.071} 	&0.207/0.118	 & 0.038/0.051 	& 0.074/0.094
     \\ 
Diff-IP3D \cite{ma2024diff}  & 0.199/0.106     &0.377/0.159	 &0.156/0.094  &0.229/0.140	  &0.049/0.061    & 0.081/0.098
   \\ 
MADiff3D \cite{ma2024madiff}  & \myblue{0.183}/\myblue{0.078}     &0.363/\myblue{0.124}	 &0.139/0.072  &0.224/\myblue{0.112}	  &0.032/\myblue{0.039}    & 0.059/0.071
   \\
\rowcolor{lightgray}
MMTwin (ours)  &\textbf{0.170}/\textbf{0.071}	  &\textbf{0.336}/\textbf{0.118}	  &\textbf{0.118}/\textbf{0.061}	  &\myblue{0.189}/\textbf{0.099}  	&\textbf{ 0.030}/\textbf{0.037 } 	&\textbf{0.050}/\textbf{0.039} \\ \bottomrule
\end{tabular}
\label{tab:compare_hand_egopat_h2o_3d}
\\ \vspace{-0.15cm}
\begin{flushleft}
\scriptsize
$^*$\,The baselines are re-evaluated according to the erratum: \url{https://github.com/oppo-us-research/USST/commit/beebdb963a702b08de3a4cf8d1ac9924b544abc4}.
\end{flushleft}
\vspace{-0.7cm}
\end{table*}

\begin{figure}[t]
  \centering
    \includegraphics[width=1\linewidth]{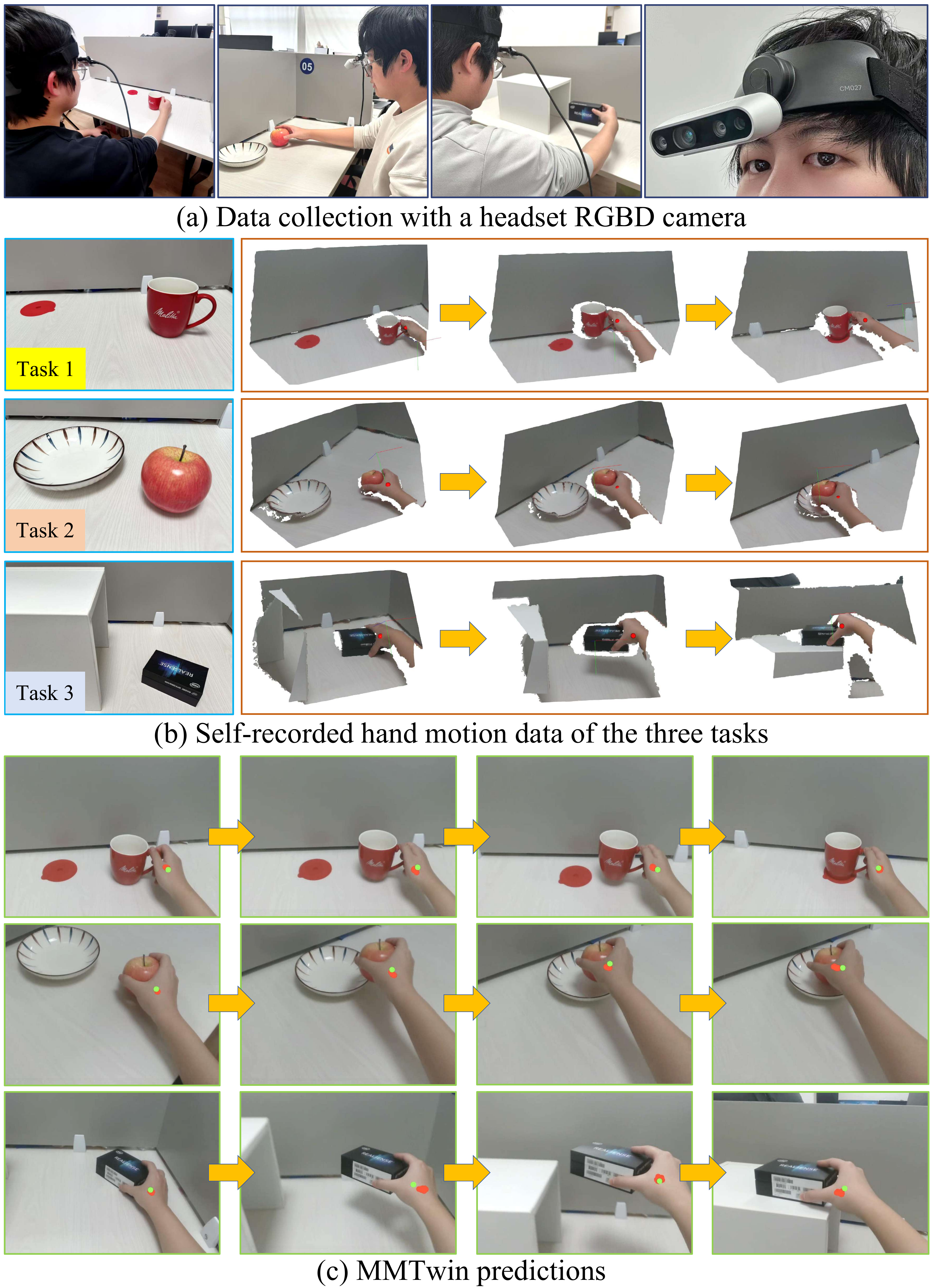}
  \caption{We use a headset RGBD camera (a) to obtain self-recorded data (b). Here we also visualize the corresponding MMTwin predictions after 10 denoising processes, projected to the image plane (c), where MMTwin predictions and ground-truth future hand waypoints are represented as red and green points respectively.}
  \label{fig:our_dataset}
  \vspace{-0.7cm}
\end{figure}

\section{Experiments}
\label{sec:exp}

\subsection{Experimental Setups}
\label{sec:setups}

\textbf{Datasets.} We evaluate our proposed MMTwin on three publicly available datasets, including EgoPAT3D-DT~\cite{li2022egocentric,bao2023uncertainty}, H2O-PT~\cite{kwon2021h2o,bao2023uncertainty}, and HOT3D-Clips~\cite{banerjee2024hot3d}, as well as our self-recorded data. Following the setups of USST \cite{bao2023uncertainty}, we use the fixed ratio 60\% by default to split the past and future sequences for both EgoPAT3D-DT and H2O-PT at 30 FPS. Each video clip in HOT3D-Clips with a duration of 5\,s is first downsampled from 30 FPS to 10 FPS, and then we also use 60\% to split past and future sequences. Note that we only adopt the Aria part of HOT3D-Clips because the other part from Quest 3 does not provide an RGB image stream. Besides, we split its official training data into the devised training and test sets for this work since the ground-truth hand annotations of the official test set are not available. Ultimately, we obtain 6356 sequences for training and 1605/2334 counterparts for testing on seen/unseen scenes on EgoPAT3D-DT, and 8203 sequences for training and 3715 counterparts for testing on H2O-PT. For HOT3D-Clips, there are randomly sampled 2732 and 300 sequences for training and testing respectively, considering both left and right hands. To further demonstrate that our method has the potential to scale up with low-cost devices for data collection, we used headset RealSense D435i to collect 1200 egocentric videos for three real-world tasks, i.e., \textit{place the cup on the coaster} (Task~1), \textit{put the apple on the plate} (Task~2), and \textit{place the box on the shelf} (Task~3) as shown in Fig.~\ref{fig:our_dataset}. For each task, 350 video clips are used for training with the other 50 clips for evaluation. Each clip is with around 5 seconds, with the first 50\% regarded as the past sequences and the latter 50\% as the future ones. We will release our self-recorded data as a new open-source HTP benchmark.

\textbf{MMTwin configuration.} We voxelize input point clouds into $20\times 20\times 20$ grids with the resolution of $0.05$\,m. The voxel patches are with the size of $27 \times 1024$.
For both twin diffusion models, we set the channel dimension of the latent features to $1024$. The total number of diffusion steps is set to $1000$, while the egomotion diffusion takes only one step to predict egomotion features for high efficiency and the HTP diffusion takes 100 steps to predict future HTP features. 
Egomotion-aware Mamba blocks of HMTM are with convolutional kernel size $d\_conv\!=\!2$, hidden state expansion $expand\!=\!1$, and hidden dimension $d\_state\!=\!16$. The number of heads in the structure-aware Transformer of HMTM is set to $n\_head=4$, and the intermediate dimension of the feed-forward layer is $d\_f\mkern-2mu f\mkern-2mu n \!=\! 2048$. 
We train MMTwin using AdamW optimizer \cite{kingma2014adam} with a learning rate of 5e-5 for 1K epochs on EgoPAT3D-DT, H2O-PT, and our self-recorded datasets, and with a learning rate of 5e-6 for 2K epochs on the HOT3D-Clips dataset. Training and inference are both operated on 2 NVIDIA A100 GPUs.

\textbf{Baseline configuration.} We select Constant Velocity Hand (CVH)~\cite{ma2024diff}, OCT~\cite{liu2022joint}, USST~\cite{bao2023uncertainty}, S-Mamba~\cite{wang2024mamba}, Diff-IP2D~\cite{ma2024diff}, and MADiff~\cite{ma2024madiff} to conduct the baselines in this work. We modify S-Mamba originally designed for general time series forecasting into our diffusion-based paradigm to predict HTP tokens. We additionally replace the 2D input and output, and the corresponding encoders and decoders with 3D counterparts in Diff-IP2D and MADiff since they were originally developed for 2D HTP tasks, obtaining the baselines Diff-IP3D and MADiff3D.

\textbf{Evaluation metrics.} Following previous works \cite{liu2020forecasting,liu2022joint,bao2023uncertainty}, we use Average Displacement Error (ADE) and Final Displacement Error (FDE) to evaluate prediction performance in both 2D and 3D spaces. The evaluation in the 3D space follows the absolute scale in meters, while we project 3D hand waypoints to the image plane and further normalize them by the image size for the evaluation in the 2D space.

\begin{figure*}
  \centering
  \includegraphics[width=1\linewidth]{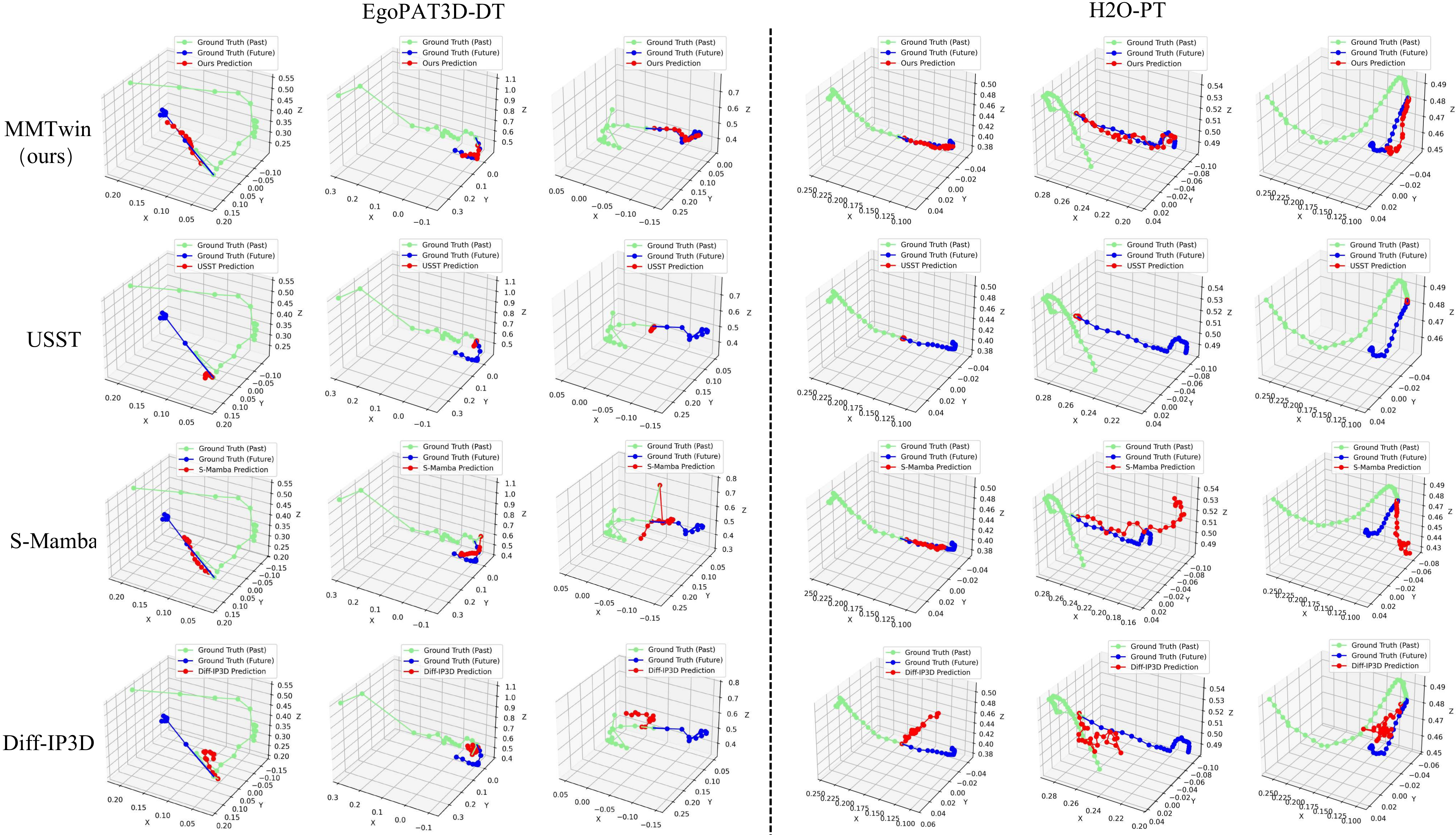}
  \caption{Visualization of predicted hand trajectories in the 3D space. We show the holistic sequence including observed past hand waypoints (green), ground-truth future ones (blue), and predicted future counterparts (red) by our MMTwin and three SOTA HTP baselines.}
  \label{fig:viz_egopat_h2o}
  \vspace{-0.2cm}
\end{figure*}

\subsection{Comparison with SOTA Approaches}
\label{sec:comparison_with_prior}
\begin{figure}
  \centering
  \includegraphics[width=1\linewidth]{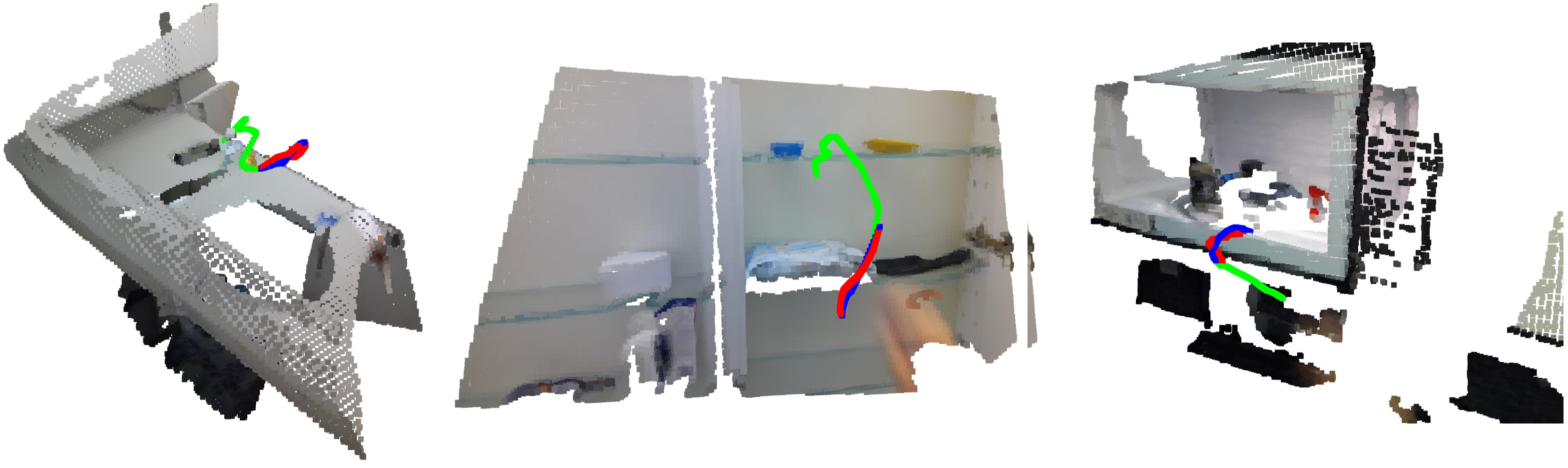}
  \vspace{-0.6cm}
  \caption{Visualization of the past hand waypoints (green), ground-truth future hand waypoints (blue), and future counterparts predicted by our MMTwin (red) with point clouds in \textit{bathroomCabinet}, \textit{bathroomCounter}, and \textit{microwave} scenes of EgoPAT3D-DT.}
  \label{fig:viz_with_points}
  \vspace{-0.7cm}
\end{figure}

We first compare our MMTwin with the selected SOTA baselines mentioned in Sec.~\ref{sec:setups} on the performance of hand trajectory prediction. 
As Tab.~\ref{tab:compare_hand_egopat_h2o_3d} shows, on the EgoPAT3D-DT and H2O-PT datasets, our proposed MMTwin achieves the best HTP performance on most metrics in 2D and 3D spaces compared to the SOTA baselines. The good HTP performance for unseen environments also demonstrates our MMTwin's solid generalization ability. We further provide visualizations of predicted hand waypoints in Fig.~\ref{fig:viz_egopat_h2o}. As can be seen, our MMTwin generates future trajectories with higher accuracy and more natural shapes. In contrast, USST tends to generate relatively short conservative trajectories, and Diff-IP3D holds higher directional uncertainties due to its model characteristics only designed for the 2D predictive tasks. Fig.~\ref{fig:viz_with_points} also illustrates the hand trajectories predicted by our MMTwin with the point clouds of exampled scenes. In addition, as depicted in Tab.~\ref{tab:compare_hand_hot3d}, our proposed MMTwin outperforms the other SOTA baselines on the HOT3D-Clips dataset, which encompasses video clips that are longer than twice the duration of the videos in EgoPAT3D-DT and H2O-PT. Because there are no point clouds available in HOT3D-Clips data, we omit the voxel patches for the structure-aware Transformer of MMTwin and replace its cross-attention with self-attention. This also demonstrates that MMTwin can still predict accurate 3D hand waypoints without valid 3D observations, which is important in some sensor-limited applications. For our self-recorded dataset in Fig.~\ref{fig:our_dataset}, Tab.~\ref{tab:compare_hand_our_data} indicates that MMTwin still outperforms the SOTA baselines even with low-cost data collection on our three tasks.

\begin{table}[t]
\setlength{\tabcolsep}{9.9pt}
\center
\renewcommand\arraystretch{0.7}
\caption{Comparison of performance on hand trajectory prediction on the HOT3D-Clips dataset in the 3D and 2D spaces. Best and secondary results are viewed in \textbf{bold black} and \myblue{blue} colors.}
\begin{tabular}{l|cc|cc}
\toprule
\multicolumn{1}{l|}{\multirow{2}{*}{Approach}}   & \multicolumn{2}{c|}{3D} & \multicolumn{2}{c}{2D}   \\ \cmidrule{2-5} 
\multicolumn{1}{c|}{}                                                                               & ADE\,$\downarrow$    & FDE\,$\downarrow$ & ADE\,$\downarrow$    & FDE\,$\downarrow$    \\ \cmidrule{1-5}    
CVH~\cite{ma2024diff}  & 1.273 & 1.358 & 0.437  & 0.443        \\
OCT~\cite{liu2022joint}  & 0.188  & 0.215  & 0.207  &  0.242       \\
USST~\cite{bao2023uncertainty}   & 0.123     &  0.157   &  \myblue{0.135}     &  0.169             \\
S-Mamba~\cite{wang2024mamba}   & \myblue{0.117}	 &  \myblue{0.132}	 & 0.136	 &  \myblue{0.162}          \\
Diff-IP3D~\cite{ma2024diff}     & 0.147 	& 0.164	  & 0.173 & 0.205  \\
MADiff3D~\cite{ma2024madiff}     & 0.120 	& 0.147	  & \myblue{0.135} & 0.165  \\
\rowcolor{lightgray}
MMTwin (ours) & \textbf{0.104}	& \textbf{0.131}  & \textbf{0.121}   & \textbf{0.155}
  \\  \bottomrule
\end{tabular}
\label{tab:compare_hand_hot3d}
\vspace{-0.2cm}
\end{table}

\begin{table}[t]
\setlength{\tabcolsep}{4pt}
\center
\renewcommand\arraystretch{0.7}
\caption{Comparison of performance on hand trajectory prediction on the self-recorded data in the 3D space. Best and secondary results are viewed in \textbf{bold black} and \myblue{blue} colors.}
\begin{tabular}{l|cc|cc|cc}
\toprule
\multicolumn{1}{l|}{\multirow{2}{*}{Approach}}   & \multicolumn{2}{c|}{Task 1} & \multicolumn{2}{c|}{Task 2}  & \multicolumn{2}{c}{Task 3}   \\ \cmidrule{2-7} 
\multicolumn{1}{c|}{}                                                                               & ADE\,$\downarrow$    & FDE\,$\downarrow$ & ADE\,$\downarrow$    & FDE\,$\downarrow$  & ADE\,$\downarrow$    & FDE\,$\downarrow$    \\ \cmidrule{1-7}    
USST~\cite{bao2023uncertainty}   & 0.102     &  0.125   &  0.109     &  0.128      & 0.103    & 0.130      \\
S-Mamba~\cite{wang2024mamba}   & \myblue{0.045}   & \myblue{0.055}     & \myblue{0.050}     & \myblue{0.072}   & \myblue{0.058} & \myblue{0.061}           \\
\rowcolor{lightgray}
MMTwin (ours) & \textbf{0.041}	& \textbf{0.052}  & \textbf{0.044}   & \textbf{0.061}   & \textbf{0.047}  & \textbf{0.053}
  \\  \bottomrule
\end{tabular}
\label{tab:compare_hand_our_data}
\vspace{-0.7cm}
\end{table}

\begin{table*}[t]
\setlength{\tabcolsep}{13pt}
\center
\renewcommand\arraystretch{0.7}
\caption{Ablation study on camera egomotion. SE(3) as egomotion represents the baseline replacing the input camera homography with 6-DOF poses. MMTwin w/o ED represents the baseline without the egomotion diffusion. Best results are viewed in \textbf{bold black}.}
\vspace{-0.2cm}
\begin{tabular}{l|cc|cc|cc}
\toprule
\multicolumn{1}{l|}{\multirow{2}{*}{Approach}}   & \multicolumn{2}{c|}{EgoPAT3D-DT (seen)} & \multicolumn{2}{c|}{EgoPAT3D-DT (unseen)}  & \multicolumn{2}{c}{H2O-PT} \\ \cmidrule{2-7} 
\multicolumn{1}{c|}{}                                                                               & ADE\,$\downarrow$    & FDE\,$\downarrow$ & ADE\,$\downarrow$   & FDE\,$\downarrow$ & ADE\,$\downarrow$   & FDE\,$\downarrow$   \\ \cmidrule{1-7}      
SE(3) as egomotion  & 0.257/0.183  & 0.446/0.244   & 0.217/0.163    & 0.308/0.216   & 0.032/0.041   & 0.063/0.077     \\ 
MMTwin w/o ED  & 0.186/0.091 & 0.363/0.139  & 0.137/0.077  & 0.231/0.120  & 0.031/0.040  & 0.053/0.045   \\ 
\rowcolor{lightgray}
MMTwin  &\textbf{0.170}/\textbf{0.071}	  &\textbf{0.336}/\textbf{0.118}	  &\textbf{0.118}/\textbf{0.061}	  &\textbf{0.189}/\textbf{0.099}  	&\textbf{0.030}/\textbf{0.037} 	&\textbf{0.050}/\textbf{0.039} \\ \midrule
Error reduction by ED  & 8.6\%/22.0\% & 7.4\%/15.1\% & 13.9\%/20.8\% & 18.2\%/17.5\% & 3.2\%/7.5\%  & 5.7\%/13.3\% \\ \bottomrule
\end{tabular}
\label{tab:abla_egomotion}
\\
\begin{flushleft}
\end{flushleft}
\vspace{-0.7cm}
\end{table*}

\begin{table*}[t]
\setlength{\tabcolsep}{12.0pt}
\center
\renewcommand\arraystretch{0.7}
\caption{Comparison of performance on hand trajectory prediction on different hybrid patterns of the EAM and SAT blocks in the hybrid Mamba-Transformer module of MMTwin. Best and secondary results are viewed in \textbf{bold black} and \myblue{blue} colors.}
\vspace{-0.2cm}
\begin{tabular}{c|l|cc|cc|cc}
\toprule
\multicolumn{1}{c|}{\multirow{2}{*}{Version}} 
 &\multicolumn{1}{l|}{\multirow{2}{*}{Hybrid pattern}}   & \multicolumn{2}{c|}{EgoPAT3D-DT (seen)} & \multicolumn{2}{c|}{EgoPAT3D-DT (unseen)}  & \multicolumn{2}{c}{H2O-PT} \\ \cmidrule{3-8} 
\multicolumn{1}{c|}{}  &\multicolumn{1}{c|}{}      & ADE\,$\downarrow$    & FDE\,$\downarrow$ & ADE\,$\downarrow$   & FDE\,$\downarrow$ & ADE\,$\downarrow$   & FDE\,$\downarrow$   \\ \cmidrule{1-8} 
1 &SAT-EAM  & 0.184/0.076	&0.353/0.117	&0.147/0.066	&0.236/0.100	 &0.033/0.042  	&0.057/0.050 \\ 
2 &EAM-SAT   &0.177/\myblue{0.072}	&0.345/\myblue{0.116}	&0.133/\myblue{0.063}	&0.217/\textbf{0.098}	&\myblue{0.031}/0.042	&0.054/0.048
 \\ 
3 &SAT-EAM-EAM  &\myblue{0.173}/0.073	 &\myblue{0.339}/\textbf{0.113}  &\myblue{0.132}/0.064 &\myblue{0.198}/\textbf{0.098}	&\textbf{0.030}/\textbf{0.037}	&\myblue{0.051}/\textbf{0.039} \\ 
4 &EAM-SAT-EAM  &0.226/0.151	 &0.402/0.207	&0.181/0.135 	&0.266/0.186  &\myblue{0.031}/\myblue{0.038}	&0.053/\myblue{0.045} \\ 
\rowcolor{lightgray}
5 &EAM-EAM-SAT  &\textbf{0.170}/\textbf{0.071}	  &\textbf{0.336}/0.118	  &\textbf{0.118}/\textbf{0.061}	  &\textbf{0.189}/\myblue{0.099}  	&\textbf{0.030}/\textbf{0.037} 	&\textbf{0.050}/\textbf{0.039} \\ \bottomrule
\end{tabular}
\label{tab:ab_hybrid_pattern}
\\
\begin{flushleft}
\end{flushleft}
\vspace{-1cm}
\end{table*}

\begin{table}[t]
\setlength{\tabcolsep}{3.3pt}
\center
\caption{Ablation study on multimodal inputs. Best results are viewed in \textbf{bold black}.}
\vspace{-0.2cm}
\renewcommand\arraystretch{0.7}
\begin{tabular}{cccc|cc|cc}
\toprule
\multicolumn{4}{c|}{Input modalities}  &\multicolumn{2}{c|}{Seen} &\multicolumn{2}{c}{Unseen}  \\ \midrule
waypoint & image & text & point cloud  & ADE\,$\downarrow$ & FDE\,$\downarrow$ & ADE\,$\downarrow$  & FDE\,$\downarrow$   \\ \midrule
\ding{51}      &       &     &         &0.178	&0.356	&0.124	&0.205  \\
 \ding{51}             & \ding{51}    &    &        &0.173	&0.350	&0.122	&0.201  \\
\ding{51}            & \ding{51}    & \ding{51}    &   &0.171 	&0.347	 &0.122	  &0.200\\ 
\ding{51}            & \ding{51}    &\ding{51}  &\ding{51}     &\textbf{0.170}	&\textbf{0.336}	&\textbf{0.118}	&\textbf{0.189}  \\ \bottomrule
\end{tabular}
\label{tab:ala_on_inputs}
\vspace{-0.8cm}
\end{table}

\subsection{Ablation Studies}
\label{sec:exp_albation}
\textbf{Camera egomotion prediction.} We first ablate the egomotion prediction by removing the egomotion diffusion. Specifically, we conduct a baseline regarding the last past camera homography as the constant egomotion in the future time horizons. Tab.~\ref{tab:abla_egomotion} presents that predicting future egomotion improves the HTP performance. This demonstrates that MMTwin decoupling the predictions of camera egomotion and hand movements understands the synergy between them within the future interaction process better. Note that there is a more significant decrease in ADE and FDE on EgoPAT3D-DT than the counterparts on H2O-PT. The reason could be that EgoPAT3D-DT holds more diverse intense head motion than H2O-PT, leading to more comprehensive supervision and a more obvious effect of egomotion prediction. 

\textbf{Camera egomotion representation.} In Tab.~\ref{tab:abla_egomotion}, we also present the HTP performance when we regard SE(3) as camera egomotion for MMTwin instead of homography. Specifically, we obtain the 6-DOF poses from visual odometry, which are embedded as egomotion features in MMTwin. As can be seen, the HTP performance drops significantly once our vanilla egomotion homography features in MMTwin are replaced with SE(3) features. The reason could be that observed hands are only encompassed within 2D image plane and camera homography matrices are more suited to representing egomotion changes entangled with hand movements.

\textbf{Multimodal inputs.} We provide an additional ablation study on multimodal inputs for MMTwin. We incrementally add past hand waypoints, RGB images, text prompt, and point clouds in model inputs. The results on EgoPAT3D-DT shown in Tab.~\ref{tab:ala_on_inputs} indicate that each input modality contributes to the ultimate HTP performance.

\textbf{Hybrid architectures.} Here we explore MMTwin performance with different hybrid patterns of Mamba and Transformer in HMTM. Due to resource limitations in possible real-world deployment, we only consider different combinations of one/two EAM blocks and one structure-aware Transformer here. We leave scaling up the respective number of Mamba and Transformer modules as our future work. As shown in Tab.~\ref{tab:ab_hybrid_pattern}, version 5 and version 3 overall predict more accurate hand waypoints than version 4. This indicates that consecutively stacked EAM blocks help to enhance hand-state modeling. Besides, version 5 and version 2 generally outperform version 3 and version 1 respectively on 3D-space evaluation metrics. That is, the posterior Transformer module leads to a more positive impact on HTP performance. The reason could be that temporal modeling achieved by EAM blocks followed by the cross-attention of Structure-Aware Transformer helps maintain the stability of HTP feature updates caused by 3D global context incorporation.

\vspace{-0.1cm}
\section{Conclusion}
\vspace{-0.1cm}
\label{sec:discuss}
In this paper, we propose novel twin diffusion models MMTwin for 3D hand trajectory prediction in egocentric views. MMTwin absorbs multimodal data including 2D RGB images, 3D point clouds, past hand waypoints, and text prompt. It concurrently predicts future camera egomotion and hand trajectories. 
Experimental results validate that MMTwin generally outperforms the SOTA baselines and shows good generalization ability to unseen environments. We hope that the paradigm of concurrently predicting camera egomotion and human body motion proposed in this work could inspire future works on human-object interaction. In the future, we will explore scaling up the hybrid patterns of Mamba and Transformer for denoising diffusion, and consider deploying the proposed method to wearable devices and robots.

\bibliographystyle{ieeetr}

\footnotesize{
\bibliography{root}}

\begin{thebibliography}{10}

\bibitem{bao2022opental}
W.~Bao, Q.~Yu, and Y.~Kong, ``Opental: Towards open set temporal action localization,'' in {\em CVPR}, pp.~2979--2989, 2022.

\bibitem{xu2023dynamic}
X.~Xu, Y.-L. Li, and C.~Lu, ``Dynamic context removal: A general training strategy for robust models on video action predictive tasks,'' {\em IJCV}, vol.~131, no.~12, pp.~3272--3288, 2023.

\bibitem{qi2024uncertainty}
Z.~Qi, S.~Wang, W.~Zhang, and Q.~Huang, ``Uncertainty-boosted robust video activity anticipation,'' {\em TPAMI}, 2024.

\bibitem{Ye_2023_ICCV}
Y.~Ye, P.~Hebbar, A.~Gupta, and S.~Tulsiani, ``Diffusion-guided reconstruction of everyday hand-object interaction clips,'' in {\em ICCV}, pp.~19717--19728, October 2023.

\bibitem{zhang2024hoidiffusion}
M.~Zhang, Y.~Fu, Z.~Ding, S.~Liu, Z.~Tu, and X.~Wang, ``Hoidiffusion: Generating realistic 3d hand-object interaction data,'' in {\em CVPR}, pp.~8521--8531, 2024.

\bibitem{zhu2023get}
Z.~Zhu and D.~Damen, ``Get a grip: Reconstructing hand-object stable grasps in egocentric videos,'' {\em arXiv preprint arXiv:2312.15719}, 2023.

\bibitem{bao2023uncertainty}
W.~Bao, L.~Chen, L.~Zeng, Z.~Li, Y.~Xu, J.~Yuan, and Y.~Kong, ``Uncertainty-aware state space transformer for egocentric 3d hand trajectory forecasting,'' in {\em ICCV}, pp.~13702--13711, 2023.

\bibitem{liu2022joint}
S.~Liu, S.~Tripathi, S.~Majumdar, and X.~Wang, ``Joint hand motion and interaction hotspots prediction from egocentric videos,'' in {\em CVPR}, pp.~3282--3292, 2022.

\bibitem{ma2024diff}
J.~Ma, J.~Xu, X.~Chen, and H.~Wang, ``Diff-ip2d: Diffusion-based hand-object interaction prediction on egocentric videos,'' {\em arXiv preprint arXiv:2405.04370}, 2024.

\bibitem{ma2024madiff}
J.~Ma, X.~Chen, W.~Bao, J.~Xu, and H.~Wang, ``Madiff: Motion-aware mamba diffusion models for hand trajectory prediction on egocentric videos,'' {\em arXiv preprint arXiv:2409.02638}, 2024.

\bibitem{li2022egocentric}
Y.~Li, Z.~Cao, A.~Liang, B.~Liang, L.~Chen, H.~Zhao, and C.~Feng, ``Egocentric prediction of action target in 3d,'' in {\em CVPR}, pp.~20971--20980, IEEE, 2022.

\bibitem{bahl2023affordances}
S.~Bahl, R.~Mendonca, L.~Chen, U.~Jain, and D.~Pathak, ``Affordances from human videos as a versatile representation for robotics,'' in {\em CVPR}, pp.~13778--13790, 2023.

\bibitem{mendonca2023structured}
R.~Mendonca, S.~Bahl, and D.~Pathak, ``Structured world models from human videos,'' {\em arXiv preprint arXiv:2308.10901}, 2023.

\bibitem{singh2024hand}
H.~G. Singh, A.~Loquercio, C.~Sferrazza, J.~Wu, H.~Qi, P.~Abbeel, and J.~Malik, ``Hand-object interaction pretraining from videos,'' {\em arXiv preprint arXiv:2409.08273}, 2024.

\bibitem{chen2024object}
Y.~Chen, C.~Wang, Y.~Yang, and C.~K. Liu, ``Object-centric dexterous manipulation from human motion data,'' {\em arXiv preprint arXiv:2411.04005}, 2024.

\bibitem{ju2025robo}
Y.~Ju, K.~Hu, G.~Zhang, G.~Zhang, M.~Jiang, and H.~Xu, ``Robo-abc: Affordance generalization beyond categories via semantic correspondence for robot manipulation,'' in {\em ECCV}, pp.~222--239, Springer, 2025.

\bibitem{liu2020forecasting}
M.~Liu, S.~Tang, Y.~Li, and J.~M. Rehg, ``Forecasting human-object interaction: joint prediction of motor attention and actions in first person video,'' in {\em ECCV}, pp.~704--721, 2020.

\bibitem{zhang2024pear}
Z.~Zhang, H.~Luo, W.~Zhai, Y.~Cao, and Y.~Kang, ``Pear: Phrase-based hand-object interaction anticipation,'' {\em arXiv preprint arXiv:2407.21510}, 2024.

\bibitem{shan2020understanding}
D.~Shan, J.~Geng, M.~Shu, and D.~F. Fouhey, ``Understanding human hands in contact at internet scale,'' in {\em CVPR}, pp.~9869--9878, 2020.

\bibitem{tang2025prompting}
B.~Tang, K.~Zhang, W.~Luo, W.~Liu, and H.~Li, ``Prompting future driven diffusion model for hand motion prediction,'' in {\em ECCV}, pp.~169--186, Springer, 2025.

\bibitem{gamage2021so}
N.~M. Gamage, D.~Ishtaweera, M.~Weigel, and A.~Withana, ``So predictable! continuous 3d hand trajectory prediction in virtual reality,'' in {\em The 34th Annual ACM Symposium on User Interface Software and Technology}, pp.~332--343, 2021.

\bibitem{lowe2004distinctive}
D.~G. Lowe, ``Distinctive image features from scale-invariant keypoints,'' {\em IJCV}, vol.~60, pp.~91--110, 2004.

\bibitem{fischler1981random}
M.~A. Fischler and R.~C. Bolles, ``Random sample consensus: a paradigm for model fitting with applications to image analysis and automated cartography,'' {\em Communications of the ACM}, vol.~24, no.~6, pp.~381--395, 1981.

\bibitem{Li_2022_CVPR}
L.~H. Li, P.~Zhang, H.~Zhang, J.~Yang, C.~Li, Y.~Zhong, L.~Wang, L.~Yuan, L.~Zhang, J.-N. Hwang, K.-W. Chang, and J.~Gao, ``Grounded language-image pre-training,'' in {\em CVPR}, pp.~10965--10975, June 2022.

\bibitem{gu2023mamba}
A.~Gu and T.~Dao, ``Mamba: Linear-time sequence modeling with selective state spaces,'' {\em arXiv preprint arXiv:2312.00752}, 2023.

\bibitem{mobile_sam}
C.~Zhang, D.~Han, Y.~Qiao, J.~U. Kim, S.-H. Bae, S.~Lee, and C.~S. Hong, ``Faster segment anything: Towards lightweight sam for mobile applications,'' {\em arXiv preprint arXiv:2306.14289}, 2023.

\bibitem{detone2016deep}
D.~DeTone, T.~Malisiewicz, and A.~Rabinovich, ``Deep image homography estimation,'' {\em arXiv preprint arXiv:1606.03798}, 2016.

\bibitem{gong2022diffuseq}
S.~Gong, M.~Li, J.~Feng, Z.~Wu, and L.~Kong, ``Diffuseq: Sequence to sequence text generation with diffusion models,'' in {\em ICLR}, 2023.

\bibitem{wang2024mamba}
Z.~Wang, F.~Kong, S.~Feng, M.~Wang, H.~Zhao, D.~Wang, and Y.~Zhang, ``Is mamba effective for time series forecasting?,'' {\em Neurocomputing}, vol.~619, p.~129178, 2025.

\bibitem{kwon2021h2o}
T.~Kwon, B.~Tekin, J.~St{\"u}hmer, F.~Bogo, and M.~Pollefeys, ``H2o: Two hands manipulating objects for first person interaction recognition,'' in {\em ICCV}, pp.~10138--10148, 2021.

\bibitem{banerjee2024hot3d}
P.~Banerjee, S.~Shkodrani, P.~Moulon, S.~Hampali, S.~Han, F.~Zhang, L.~Zhang, J.~Fountain, E.~Miller, S.~Basol, {\em et~al.}, ``Hot3d: Hand and object tracking in 3d from egocentric multi-view videos,'' {\em arXiv preprint arXiv:2411.19167}, 2024.

\bibitem{kingma2014adam}
D.~P. Kingma and J.~Ba, ``Adam: A method for stochastic optimization,'' {\em arXiv preprint arXiv:1412.6980}, 2014.

\end{thebibliography}

\end{document}